\documentclass[letterpaper]{article} % DO NOT CHANGE THIS
\usepackage{aaai24}  % DO NOT CHANGE THIS
\usepackage{times}  % DO NOT CHANGE THIS
\usepackage{helvet}  % DO NOT CHANGE THIS
\usepackage{courier}  % DO NOT CHANGE THIS
\usepackage[hyphens]{url}  % DO NOT CHANGE THIS
\usepackage{graphicx} % DO NOT CHANGE THIS
\usepackage{natbib}  % DO NOT CHANGE THIS AND DO NOT ADD ANY OPTIONS TO IT
\usepackage{caption} % DO NOT CHANGE THIS AND DO NOT ADD ANY OPTIONS TO IT
\usepackage{algorithm}
\usepackage{algorithmic}
\usepackage{amsfonts}
\usepackage[dvipsnames]{xcolor}
\usepackage{paralist}
\usepackage{multirow}
\usepackage{makecell}
\usepackage{amsmath}
\usepackage{soul}

\usepackage{newfloat}
\usepackage{listings}
\DeclareCaptionStyle{ruled}{labelfont=normalfont,labelsep=colon,strut=off} % DO NOT CHANGE THIS
\floatstyle{ruled}
\newfloat{listing}{tb}{lst}{}
\floatname{listing}{Listing}
\def\eg{\emph{e.g.}}

\def\ie{\textit{i.e.}}

\newcommand{\bsb}[1]{\boldsymbol{#1}}

\newcommand{\abc}[1]{#1}
\newcommand{\abcn}[1]{#1}
\newcommand{\NOTE}[1]{}
\newcommand{\lw}[1]{#1}
\newcommand{\lwn}[1]{#1}

\newcommand{\ANS}[1]{}
\newcommand{\CUT}[1]{}

\title{A Fixed-Point Approach to Unified Prompt-Based Counting}
\author{
    Wei Lin, ~~Antoni B. Chan
}
\affiliations{
    Department of Computer Science, City University of Hong Kong\\
    Tat Chee 83, Kowloon Tong, Hong Kong SAR, China\\
    elonlin24@gmail.com, ~~abchan@cityu.edu.hk
}

\usepackage{bibentry}
\begin{document}

\maketitle

\begin{abstract}
	Existing class-agnostic counting models typically rely on a single type of prompt, e.g., box annotations. This paper aims to establish a comprehensive prompt-based counting framework capable of generating density maps for concerned objects indicated by various prompt types, such as box, point, and text. To achieve this goal, we begin by converting prompts from different modalities into prompt masks without requiring training. These masks are then integrated into a class-agnostic counting methodology for predicting density maps. Furthermore, we introduce a fixed-point inference along with an associated loss function to improve counting accuracy, all without introducing new parameters. The effectiveness of this method is substantiated both theoretically and experimentally. Additionally, a contrastive training scheme is implemented to mitigate dataset bias inherent in current class-agnostic counting datasets, a strategy whose effectiveness is confirmed by our ablation study. Our model excels in prominent class-agnostic datasets and exhibits superior performance in cross-dataset adaptation tasks.
\end{abstract}

\section{Introduction}

Visual counting has been a longstanding topic of interest, driven by the demands of industrial intelligence. However, the majority of existing methods concentrate on counting specific object categories, such as pedestrians~\cite{nwpu, otm, gcf}, animals~\cite{cwild}, or cars~\cite{ccar}. This limitation hampers their applicability to unseen categories and weakens their transferability. In contrast, numerous commercial and agricultural scenarios necessitate the counting of diverse objects, such as goods on a shelf~\cite{cpack}, various crops in farmland~\cite{ccorn, cberry}, or buildings in remote sensing images~\cite{cremote, cbuild}. To tackle these scenarios, there exists a requirement for class-agnostic models capable of counting objects of any type.

\begin{figure}
	\centering
	\includegraphics[width=0.47\textwidth]{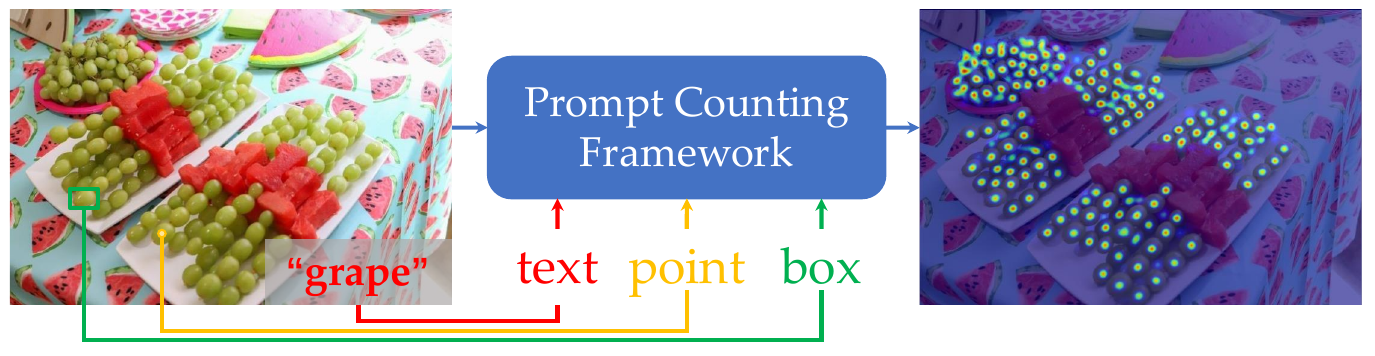}
	\caption{Overview of prompt-based counting. This model takes prompts in various modalities, \eg, box, point, or text annotation to indicate the object of interest, and then predicts the distribution and count accordingly.}
	\label{fig:overview}
\end{figure}

Recent class-agnostic counting models have primarily focused on a box-guided pipeline~\cite{gmn, fsc147, bmnet, countr, spdcn}. These methods utilize three box annotations to indicate the object of interest and subsequently predict a density map to illustrate its distribution and count. Moreover, there are also works that develop models for counting based on text prompts~\cite{zsoc, clipcount}. To establish a connection between vision and language, a language-image pre-training model,  CLIP~\cite{clip}, is employed to align text and visual embeddings within the same space. Typically, both text and visual encoders remain frozen to retain the zero-shot capability in CLIP, while the density predictor is trained within few-shot counting datasets for task adaptation. However, there are no related works combining these prompts to establish a unified prompt-based counting framework, which can use various types of prompts to predict density maps within only one model. To address this issue, we propose building a unified prompt-based counting framework, considering three types of prompts: box, point, and text, as displayed in Figure~\ref{fig:overview}. 

Normally, the prompts are represented as a tokens, \ie, feature vectors, in class-agnostic counting~\cite{bmnet, countr, spdcn}. The first problem we need to address is how to translate these prompts from different modalities into the same representation. For boxes and points, we can directly aggregate information from the corresponding regions. However, since text prompts are not visual cues, it cannot be straightforwardly mapped into the visual feature space. Inspired by MaskCLIP~\cite{maskclip}, we utilize the frozen CLIP to generate a text prompt mask that is consistent with box and point prompts. 
The key principle of this transformation is that the value embeddings in the last layer of the visual encoder in CLIP contain local semantic features that can be mapped into the language space. Thus, we can directly measure the similarity between the text feature and local value embeddings to obtain a prompt mask similar to box and point prompts. This approach eliminates the need for a training step to establish the connection between text and image, allowing the training to focus on density prediction, improving performance. 
This contrasts with previous text-guided zero-shot counting methods~\cite{zsoc, clipcount}, \lw{which needs to randomly crop numerous patches from the input images and then proceed to compare them with the text feature in order to select appropriate tokens}.

After prompt masks are obtained freely, we employ cross attention to generate density features and utilize a convolutional-based module to make the final predictions. When applying the aforementioned framework to tackle class-agnostic counting, two issues emerge that need to be addressed: (1) The model's robustness is lacking, leading to poor performance during inference; (2) The training dataset FSC-147~\cite{fsc147} exhibits a bias where most images contain only one type of object. This bias causes the model to count the \emph{salient} object instead of the one indicated by the prompt, especially when the prompt is noisy. 

To address the robustness, we introduce a fixed-point % \NOTE{unclear..fixed point what? training scheme?}
\lw{inference scheme and its corresponding loss function} into our model to enhance its performance without introducing new parameters. Specifically, we observe that the predicted density map can also be considered as a prompt mask  \abcn{to compute the prompt token}, %to perform counting, 
establishing a fixed point %\NOTE{fixed point what? criteria?} 
\lw{solution of the prompt-based counting function}
% \ANS{Here, I want to express that the density map is a fixed point of the prompt-based counting, as it satisfies (\ref{eq:fixpf}),  which corresponds to the mathematical definition of a fixed point.}
that converges to a consistent prediction. The density predictor possesses a natural recurrent structure for refining the density map. However, recurrent training %a recurrent model proves 
is challenging due to the continuous accumulation of gradients. To overcome this issue, we design a fixed-point loss based on implicit differentiation~\cite{imdiff} and bi-level optimization~\cite{bilopt} to optimize parameters effectively. Experimental results demonstrate that this design can \abc{converge the model to} %identify 
a favorable parameter space and result in lower estimation errors compared to a model without it.

To address the dataset bias, we employ a contrastive training scheme to train the counting model. For each training sample (positive), we randomly select an image from the dataset as its contrasting sample (negative). While a token is aggregated according to prompt masks from the image features, we compute two density maps: one for the positive sample and another for the negative sample. Ideally, the density map for the positive sample should closely align with its ground truth, while the density map for the negative sample should resemble an all-zero density map. This approach enables the trained model to make accurate predictions even when multiple objects are present within a single image.

In summary, this paper makes three contributions:
\begin{compactenum}
	\item We propose a unified prompt-based class-agnostic counting framework to count objects indicated by boxes, points, and texts. These prompts from different modalities are transformed into semantic masks and then counted accordingly.
	\item To improve robustness, we propose a fixed-point inference and loss function to train a recurrent structure in the counting framework. This is based on the finding that the predicted density map could also be regarded as a prompt to generate a token for counting objects, creating a fixed point and a recurrent scheme to refine the density map.
	\item Addressing dataset bias, we advocate contrastive training of the prompt counting model. The model should predict a density map close to the ground truth for positive samples and an all-zero map for the negative samples.
\end{compactenum}

\section{Related Works}

Visual object counting has been extensively studied in the literature, often concentrating on counting specific objects like pedestrians~\cite{ppcm, mcnn, crosc}, vehicles~\cite{ccar,carpk}, cells~\cite{cellc,cellc2}, crops~\cite{berryc,treec}, and animals~\cite{wildc,sheepc, fishc}. While these models exhibit impressive performance, their applicability is constrained by their inability to generalize to unfamiliar object types on which they were not trained. This limitation has prompted the need for class-agnostic counting. Recent advancements in this field have delved into both box-guided and text-guided approaches.

\subsection{Box-Guided Object Counting}

Class-agnostic counting with exemplars is commonly referred to as few-shot counting~\cite{gmn, cafc, fsc147}. In GMN~\cite{gmn}, self-similarity computations are utilized to identify similar patches to an exemplar patch that indicates what should be counted, highlighting them in the predicted heat map. CFOCNet~\cite{cafc} employs several reference images to denote the object of interest and then performs counting on a query image. Additionally, \citeauthor{fsc147} introduced a class-agnostic counting dataset, FSC-147, along with a baseline model, FamNet, where given exemplars function as convolution kernels to locate objects of a specified type within the image~\cite{fsc147}.

Following FSC-147, several box-guided class-agnostic counting approaches have been proposed for general object counting~\cite{bmnet, fcad, rcac, spdcn, countr}. Most of these approaches conceratrate on refining the matching strategy between the given exemplars and query images. For instance, BMNet~\cite{bmnet} employs a dynamic similarity metric and a scale embedding to enhance the matching score between exemplars and ground truth. RCAC~\cite{rcac} introduces feature augmentation and an edge matching module to handle intra-class diversity. CounTR~\cite{countr} corporates a transformer-based architecture, leveraging the cross-attention capabilities of transformers for matching tasks. In contrast, SPDCN~\cite{spdcn} focuses on extracting robust features for matching by integrating exemplar information into the network backbone. Counting-DETR~\cite{fcad} explores the amalgamation of few-shot counting and detection, treating objects as points and producing not only their locations but also their sizes. One limitation of box-guided approaches is that they require additional effort to annotate what should be counted using box label for each image.

\subsection{Object Counting With Text Prompts}

Counting objects according to given text involves taking an image and a text as input, and then predicting the count of the object indicated by the text within the image. ZSOC~\cite{zsoc} proposes a method using a conditional VAE to transform text cues into visual features, and subsequently utilizes a clustering method to aggregate corresponding object features in the given image. CLIP-C~\cite{clipcount} employs the visual-language pre-trained model CLIP~\cite{clip} to establish a connection between text and image. %The CLIP model remains frozen both during training and inference stages.
An intra-image contrastive and a hierarchical text-patch interaction module are designed to generate final density maps. Both models achieve outstanding performance on the %publicly available 
class-agnostic counting dataset.

In contrast to the above works that focus solely on either box or text as prompts, our proposed unified prompt-based counting framework can count objects with three types of prompts using just one model. A concurrent work, TFPOC~\cite{tfpoc}, presents a training-free counting approach based on both CLIP~\cite{clip} and SAM~\cite{sam}, also addressing this task. However, it falls short in handling the challenges poses by occlusion and blur in dense scenes, as its counting ability is rooted in instance segmentation rather than density prediction. \abc{Thus, as demonstrated in the experiments, our proposed method has lower counting error than TFPOC.}

\section{Method}

\begin{figure}
	\centering
	\includegraphics[width=0.48\textwidth]{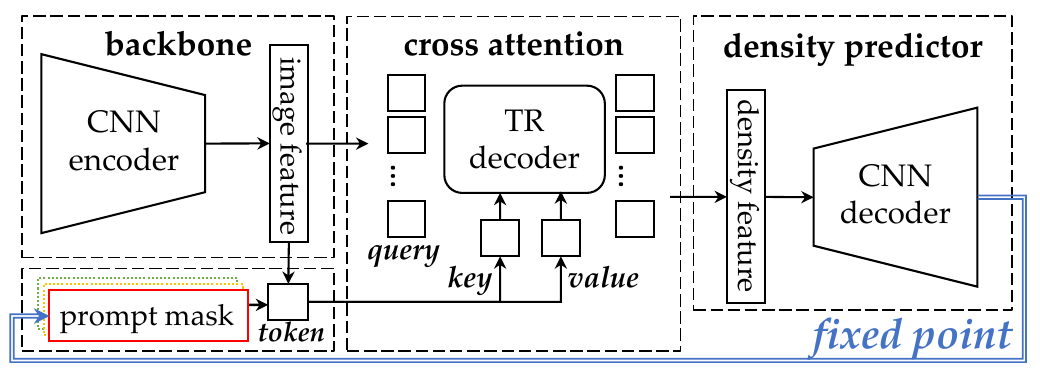}
	\caption{Our unified prompt-based counting framework. A CNN encoder generates image features, and a token is aggregated based on the provided prompt mask. Next, cross-attention is applied to generate density features, which are then decoded to produce the density map. Importantly, the density map can also be viewed as a prompt mask, implying the existence of a fixed point \abc{solution}. This fixed point enables the utilization of a loop to enhance the output.}
	\label{fig:model}
\end{figure}

In this section, we present our token-based prompt counting framework. The structure of the proposed prompt counting framework is illustrated in Figure~\ref{fig:model}. It takes an image and a prompt mask as input, and then computes the similarity between the prompt and image features, which helps filter unrelated parts in the image. The final density map is predicted by a CNN decoder to represent the distribution and count of the concerned object.

% In the subsequent parts, we first describe how to transform prompts from various modalities into a unified representation, with a particular focus on text prompts. Following that, we introduce the proposed recurrent inference and the fixed-point loss to effectively estimate density maps without altering the model structure. Specifically, a fixed point is observed within our framework when the prediction is considered as a prompt mask, prompting use to refine the density map by replacing the initial mask with the prediction during inference. The fixed-point loss is designed to address the issue of gradient explosion during training of the recurrent structure. Furthermore, we introduce the contrastive training scheme to mitigate the dataset bias within the training dataset, FSC-147~\cite{fsc147}.

\subsection{Prompt Mask}

Within our framework, the prompt could take the form of a box or point, indicating the location of an instance, or a piece of text describing what should be counted, as depicted in Figure~\ref{fig:overview}. However, the text prompt differs from the former two, as it cannot be directly transformed into visual cues. Hence, the initial challenge we must address is how to establish a uniform representation for these prompts, even when they take various modalities, \ie, box, point, and text.

The object of interest in an image can typically represented in two ways: a) by an attention mask $\bsb{m}$ that highlights regions occupied by the object; b) by a token $\bsb{t}$ that represents the object's semantic feature. 
Given an input image feature map $\bsb{F}$, the relationships among these  are:
\begin{equation}
	\bsb{t} = \tfrac{\bsb{F}^{\top}\bsb{m}}{\|\bsb{m}\|_1}, ~~ \bsb{m} \in \mathbb{R}_{+}^{N}, ~~ \bsb{F} \in \mathbb{R}^{N \times C}, \label{eq:m2t}
\end{equation}
%\NOTE{use subscript "+" to indicate non-negative}\ANS{OK.}
where $N= h \times w$ is the size of the flattened feature map, $C$ is the dimension of features, and $\|\cdot\|_1$ donates the \texttt{sum} operation, as the elements in $\bsb{m}$ are non-negative. 
% In the following parts, we detail the process of generating a prompt mask within our unified framework without  training.
For box and point prompts, $\bsb{m}$ can be formulated as a mask in which the labeled region or pixel is set to 1 and the other pixels are set to 0. 

For text prompts, we utilize CLIP~\cite{clip} to generate an prompt mask inspired by MaskCLIP~\cite{maskclip}. Specifically, the matching score between a pre-defined text feature $\bsb{f}_{T} \in \mathbb{R}^C$ and a given visual feature $\bsb{f}_{I} \in \mathbb{R}^{C}$ in CLIP is computed using cosine distance:
\begin{equation}
	\mathcal{S}_T(\bsb{f}_I) = \texttt{cosine}(\bsb{f}_T, \bsb{f}_I) =  \frac{\bsb{f}_{T}^{\top}\bsb{f}_{I}}{\|\bsb{f}_{T}\|\|\bsb{f}_{I}\|}. \label{eq:cos}
\end{equation}
Looking closely at the last layer of the visual %\NOTE{vision?}
\lw{encoder in CLIP},
% \ANS{I mean the visual encoder in CLIP, not vision transformer}, 
the visual cue can be formulated as:
\begin{align}
	\bsb{f}_{I} &=\mathcal{H}\left(\sum\nolimits_i w_i \bsb{v}_i\right) = \sum\nolimits_i w_i\mathcal{H}(\bsb{v}_i). \label{eq:clipv}	
\end{align}
Here $w_i = \texttt{softmax}\left({\bsb{q}^\top \bsb{k}_i}/{\sqrt{C}}\right)$ measures the saliency level of the $i$-th region. $\bsb{q}$ is the \textit{query} of the class embedding, while $\bsb{k}_i$ and $\bsb{v}_i$ represent the \textit{key} and \textit{value} embeddings at spatial location $i$. The equality in (\ref{eq:clipv}) holds because $\mathcal{H}$ is a linear projection. We can then form the approximation: 
\begin{equation}
	\mathcal{S}_T(\bsb{f}_I) = \mathcal{S}_T(\sum\nolimits_i w_i\mathcal{H}(\bsb{v}_i)) \approx \sum\nolimits_i w_i \mathcal{S}_T(\mathcal{H}(\bsb{v}_i)). \label{eq:mclip}
\end{equation}
Specifically, by taking the first-order Taylor approximation of $\mathcal{S}_T(\mathcal{H}(\bsb{v}))$ around $\bsb{f}_I$:
\begin{equation}
	\mathcal{S}_T(\mathcal{H}(\bsb{v})) \approx \mathcal{S}_T(\bsb{f}_I) - \nabla \mathcal{S}_T(\bsb{f}_I)^{\top}(\mathcal{H}(\bsb{v}) - \bsb{f}_I). \label{eq:taylor}
\end{equation}
Substituting (\ref{eq:taylor}) into the RHS of (\ref{eq:mclip}), we have
\begin{equation}
	\begin{aligned}
		&\sum\nolimits_i w_i\mathcal{S}_T(\mathcal{H}(\bsb{v}_i)) \\
		&\quad\approx \sum\nolimits_i w_i \left[\mathcal{S}_T(\bsb{f}_I)
		-\nabla\mathcal{S}_T(\bsb{f}_I)^{\top}(\mathcal{H}(\bsb{v}_i) - \bsb{f}_I)\right] \\
		&\quad= \mathbf{S}_T(\bsb{f}_I).
	\end{aligned}
\end{equation}
Thus (\ref{eq:mclip}) is proved. With this approximation in mind, it become reasonable to interpret  $\mathcal{S}_T(\mathcal{H}(\bsb{v}_i))$ as measuring the matching score between the text feature $\bsb{f}_T$ and local visual feature at the $i$-th position. This interpretation holds true because $w_i$ functions as a weight for aggregating salient information from the entire image.

\begin{figure}
	\centering
	\includegraphics[width=0.47\textwidth]{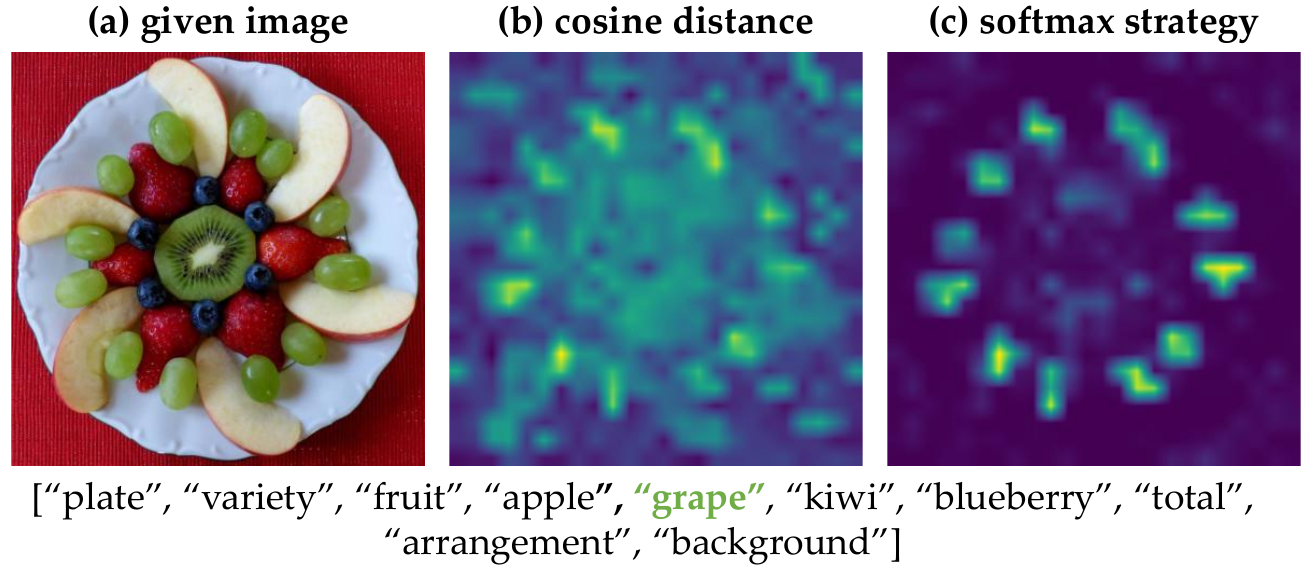}
	\caption{Comparison between \texttt{cosine} distance in (\ref{eq:cosmask}) and \texttt{softmax} strategy in (\ref{eq:sfxmask}) on text-prompt mask generation. The list shows the extracted concept dictionary via LLaMA-Adapter V2 and Spacy.
		\lw{The green text represents the user-provided text prompt, whose index is $k$ in (\ref{eq:sfxmask}).}}
	\label{fig:clip}
\end{figure}

Based on the analysis above, we derive a text prompt mask using the representation of $\texttt{cosine}$ distance:
\begin{align}
	\bsb{m} = [\texttt{cosine}\left(\bsb{f}_T, \mathcal{H}(\bsb{v}_i)\right)]_{i}, i \in \{1,\ldots,N\}. \label{eq:cosmask}
\end{align}
However, we observe that the mask generated via (\ref{eq:cosmask}) contains much noise caused by objects close to the target object in the CLIP feature space, as depicted in Figure~\ref{fig:clip}(b). To address this issue, we require a \emph{concept dictionary} that lists object categories present in the input image. This allows us to apply \texttt{softmax} to reduce the influence of background objects, similar to the approach adopted by CLIP in zero-shot learning~\cite{clip}.

To \abc{obtain the concept dictionary}, we employ an image caption model, LLaMA-Adapter~V2~\cite{llamav2}, to generate a detailed description of the input image.
% \NOTE{it is for each image separately, or you generate the dictionary from the training images}\ANS{The concept dictionary is tailored to each individual sample. Each image has its unique concept dictionary.}. 
We then utilize Spacy~\cite{spacy2} to identify all nouns and construct the concept dictionary. \lw{Note that 
	each image has its own concept dictionary.} 
%the concept dictionary is tailored to each individual sample, with each image processing its unique concept dictionary.} 
Next, we can generate the prompt mask using (\ref{eq:mclip}) and the \texttt{softmax} strategy from CLIP:
%\begin{align}
%	\bsb{m} = \left[\texttt{softmax}\left[\tau \mathcal{S}_{T_j}(\mathcal{H}(\bsb{v}_i))\right]_{[k]} \right]_i, i\in\{1,\ldots,N\}, \label{eq:sfxmask}
%\end{align}
\lwn{\begin{align}
		\bsb{m} = \left[ \tfrac{\exp\left(\tau\mathcal{S}_{T_k}(\mathcal{H}(\bsb{v}_i))\right)}{\sum_{j=1}^L\exp\left(\tau\mathcal{S}_{T_j}(\mathcal{H}(\bsb{v}_i))\right)} \right]_i, i \in \{1,\ldots,N\}, \label{eq:sfxmask}
	\end{align}
	where $\tau$ is the temperature, $T_j$ denotes the $j$-th text in the concept dictionary with a length of $L$, and $T_k$ represents the user-provided text in the concept dictionary.}
%\NOTE{the notation in (8) is a bit confusing. Is the softmax  over a vector where each element is from varying "j"?  is the "k" index applied to the resulting softmax vector?}\ANS{Yes, you are correct. Each pixel $\bsb{v}_i$ interacts with every text $T_j$ in the dictionary via $\mathcal{S}_{T_j}(\mathcal{H}(\bsb{v}_i))$, which creates a matrix $\bsb{M} \in \mathbb{R}^{N\times L}$. Softmax is applied to each row. Subsequently, the $[k]$-th column is employed as the mask $\bsb{m}$ in the origional (\ref{eq:sfxmask}). Here I have rewritten and provided a detailed formulation of softmax, aiming for greater clarity.}
An example is  in Figure~\ref{fig:clip}.

\abcn{Finally, we note that our prompt framework is flexible and can also handle other types of prompts, such as instance-masks, as well as multi-prompts (e.g., both text and box are provided). More details are provided in the supplementary.}

\subsection{Fixed-Point Inference and Loss Function}

Given the image feature $\bsb{F}$ and prompt mask $\bsb{m}$, a prompt token $\bsb{t}$ is computed according to (\ref{eq:m2t}). The final density map $\bsb{d}$ can be predicted by a function $\mathcal{D}$ with parameters $\theta$:
\begin{equation}
	\bsb{d} = \mathcal{D}_{\theta}\left(\bsb{F}, ~ \bsb{t}\right), \label{eq:tf2d}
\end{equation}
where $\mathcal{D}_{\theta}$ comprises the cross-attention module and density predictor in Figure~\ref{fig:model}.
With the predicted $\bsb{d}$ and ground truth $\bsb{d}'$, the counting model can be optimized through the L2 loss:
\begin{equation}
	\mathcal{L}(\bsb{d}, ~\bsb{d}') = \|\bsb{d} - \bsb{d}'\|^2, \label{eq:mse}
\end{equation}
which is the conventional design for class-agnostic counting.

Here we design a more effective method without altering the network structure by leveraging fixed-point theories~\cite{fixpot, diffix}. Specifically, we identify that the predicted density map $\bsb{d}$ can also function as a prompt mask to generate the semantic token, thus establishing a fixed-point iteration:
\begin{align}
	\bsb{d}^* = \mathcal{F}(\bsb{d}^*, {\Theta}) =  \mathcal{D}_{\theta}\left(\bsb{F}, ~
	\tfrac{\bsb{F}^{\top}\bsb{d}^{*}}{\|\bsb{d}^{*}\|_1}\right), \label{eq:fixpf}
\end{align}
where $\mathcal{F}(\bsb{d}^*, \Theta)$ represents the fixed-point function with respect to the fixed-point $\bsb{d}^*$, and ${\Theta} = [\bsb{F}, \theta]$ \lw{are the weights of $\mathcal{F}$}.
% \NOTE{unclear what non-renewable and renewable means in this context}. (\ref{eq:fixpf})\ANS{Here, I want to convey that $\Theta$ remains unchanged during each iteration, while $\bsb{d}^*$ is updated after each iteration until convergence. I'd like to say $\Theta$ is the parameters of $\mathcal{F}$ to avoid ambiguity.} 
(\ref{eq:fixpf}) inspires us to perform (\ref{eq:fixpro}) to refine the density map $\bsb{d}$ in a loop for improved results, initialized with $\bsb{d}^{(0)} = {\bsb{m}}$:
\begin{align}
	\bsb{d}^{(t+1)} = \mathcal{F}(\bsb{d}^{(t)}, \Theta), \quad t \in\{ 0, 1, \ldots, T-1\}, \label{eq:fixpro}
\end{align}
where $T$ represents the number of refinement iterations.

The challenge lies in the fact is that the recurrent algorithm  can lead to training instability. To address this, we introduce a loss function using implicit differentiation~\cite{imdiff} and bi-level optimization~\cite{bilopt}. We compute the gradient of $\bsb{d}^*$ with respect to $\Theta$ in (\ref{eq:fixpf}) utilizing the implicit function theorem~\cite{diffix}:
\begin{align}
\tfrac{\partial \bsb{d}^*}{\partial \Theta} &= \tfrac{\partial \mathcal{F}(\bsb{d}^*, \bsb{F})}{\partial \Theta} + \tfrac{\partial \mathcal{F}(\bsb{d}^*, \bsb{F})}{\partial \bsb{d}^*} \tfrac{\partial \bsb{d}^*}{\partial \Theta}, \\
\Rightarrow ~~ \tfrac{\partial \bsb{d}^*}{\partial \Theta} &= \left[\bsb{I} - \tfrac{\partial \mathcal{F}(\bsb{d}^*, \bsb{F})}{\partial \bsb{d}^*}\right]^{-1} \tfrac{\partial \mathcal{F}(\bsb{d}^*, \bsb{F})}{\partial \Theta} \label{eq:imf}
\end{align}
The inverse item of (\ref{eq:imf}) can be expressed using Neumann series:
\begin{align}
\left[\bsb{I} - \tfrac{\partial \mathcal{F}(\bsb{d}^*, \bsb{F})} {\partial \bsb{d}^*}\right]^{-1} = \sum\nolimits_{k = 0}^{K=\infty}\left[\tfrac{\partial \mathcal{F}(\bsb{d}^*, \bsb{F})} {\partial \bsb{d}^*}\right]^{k}. \label{eq:nms}
\end{align}
Thus, the gradient of $\mathcal{L}$ with respect to $\Theta$ is:
\begin{align}
\tfrac{\partial \mathcal{L}}{\partial \Theta} &= \tfrac{\partial \mathcal{L}}{\partial \bsb{d}^*}\left[\bsb{I} - \tfrac{\partial \mathcal{F}(\bsb{d}^*, \bsb{F})} {\partial \bsb{d}^*} \right]^{-1} \tfrac{\partial \mathcal{F}(\bsb{d}^*, \bsb{F})}{\partial \Theta} \\
&= \sum\nolimits_{k = 0}^{K=\infty}\tfrac{\partial \mathcal{L}}{\partial \bsb{d}^*}\left[\tfrac{\partial \mathcal{F}(\bsb{d}^*, \bsb{F})} {\partial \bsb{d}^*}\right]^{k}\tfrac{\partial \mathcal{F}(\bsb{d}^*, \bsb{F})}{\partial \Theta}. \label{eq:gradlt}
\end{align}

Subsequently, we take the first-order approximation for practical computation of (\ref{eq:gradlt}):
\begin{align}
\tfrac{\partial \mathcal{L}}{\partial \Theta} \approx \tfrac{\partial \mathcal{L}}{\partial \bsb{d}^*} \tfrac{\partial \mathcal{F}(\bsb{d}^*, \bsb{F})} {\partial \Theta} = \tfrac{\partial \mathcal{L}(\mathcal{F}(\bsb{d}^*, \Theta), \bsb{d}')}{\partial \Theta}, \label{eq:gradt}
\end{align}
which implies that the gradient of $\Theta$ after iterations can be approximated by only computing the last iteration on $\bsb{d}^*$ instead of employing detailedly backpropagation within the recurrent structure.

In the ideal scenario, the last prediction $\bsb{d}^{(T)}$ in (\ref{eq:fixpro}), the result of infinite iteration $\bsb{d}^{(\infty)}$, the fixed point $\bsb{d}^*$, and the ground truth $\bsb{d}'$ are extremely close to each other:
\CUT{\footnote{$\|\bsb{d}^* - \bsb{d}'\|_2 \ll \|\bsb{d}^{(T)} - \bsb{d}^{(\infty)}\|_2 < \epsilon$ in the manifold of $\bsb{d}$, where $\epsilon > 0$.}:}
\begin{align}
	\bsb{d}^{(T)} \approx \bsb{d}^{(\infty)} = \bsb{d}^* \simeq \bsb{d}'. \label{eq:dd}
\end{align}
As a result, we replace $\bsb{d}^*$ with $\bsb{d}'$ in (\ref{eq:gradt}) and formulate the following loss to incorporate (\ref{eq:gradt}) into training,% into the computation graph:
\begin{align}
	\mathcal{L}_{\infty} = \mathcal{L}(\mathcal{F}(\bsb{d}', \Theta), ~\bsb{d}'). \label{eq:lossi}
\end{align}
Furthermore, to minimize the error between $\bsb{d}^{(T)}$ and $\bsb{d}^{(\infty)}$, we also utilize the following loss during training:
\begin{align}
	\mathcal{L}_T = \mathcal{L}(\bsb{d}^{(T)}, ~\mathcal{F}(\bsb{d}', \Theta)). \label{eq:losst}
\end{align}
By assuming $\bsb{d}^{(\infty)} = \mathcal{F}(\bsb{d}', \Theta)$ and combining (\ref{eq:lossi}) and (\ref{eq:losst}), we formulate the proposed fixed-point loss function as:
\begin{align}
	\hat{\mathcal{L}} = \mathcal{L}_{\infty} + \mathcal{L}_{T} = \mathcal{L}(\bsb{d}^{(T)}, \bsb{d}^{(\infty)}) + \mathcal{L}(\bsb{d}^{(\infty)}, \bsb{d}')
	. \label{eq:floss}
\end{align}
Experimental results demonstrate that the optimal performance is achieved when $T$ is set to 2.
\subsection{Contrastive Training Strategy}

FSC-147~\cite{fsc147} serves as the main dataset for training a class-agnostic counting model. However, a dataset bias exists: for most training images, only one type of object exists in the training image. This bias may cause the model training to take a shortcut and count the salient object rather than the objects indicated by the prompt, especially in case where the provided prompts contain much noise. To address this issue, we employ a contrastive training strategy. The key principle is that a prompt token in one image should correspond to an all-zero density map in another image \lw{if they do not contain the same type of objects.} 
%\NOTE{independence is not the correct word. The images are statistically independent already. I think you mean they don't contain the same object.}. \ANS{Yes, I mean they don't cibtaub the same type of object.}

A training sample is represented as $(\bsb{F}, \bsb{m}, \bsb{d}')$, where $\bsb{F}$ denotes image features, $\bsb{m}$ and $\bsb{d}'$ represent the prompt mask and the ground truth, respectively. To generate samples for contrastive training, we concatenate the $i$-th image with the $j$-th one in a batch, creating a new training sample, denoted as $\bsb{F}_{ij} = [\bsb{F}_i, \bsb{F}_j]$. Furthermore, we concatenate $\bsb{m}_i$ and $\bsb{d}'_i$ with two all-zero tensors, $\bsb{0}_{\bsb{m}_j}$ and $\bsb{0}_{\bsb{d}'_j}$ respectively, whose shape comes from the subscript, to formulate the new prompt mask and learning target, $\bsb{m}_{ij} = [\bsb{m}_i, \bsb{0}_{\bsb{m}_j}]$ and $\bsb{d}'_{ij} = [\bsb{d}'_i, \bsb{0}_{\bsb{d}'_j}]$. This approach allows us to define the fixed-point loss within contrastive training as follows:
\begin{align}
	\hat{\mathcal{L}}_{ij} = \mathcal{L}(\bsb{d}^{(T)}_{ij}, \bsb{d}^{(\infty)}_{ij}) + \mathcal{L}(\bsb{d}^{(\infty)}_{ij}, \bsb{d}'_{ij}). \label{eq:closs}
\end{align}
In this context, the definition of $\bsb{d}^{(t)}_{ij}$ is consistent with (\ref{eq:fixpro}):
\begin{align}
	\bsb{d}^{(t+1)}_{ij} = \mathcal{F}(\bsb{d}^{(t)}_{ij}, \Theta) \quad\text{and}\quad \bsb{d}^{(0)}_{ij} = \bsb{m}_{ij}.
\end{align}
Through this approach, we enforce the given prompt to predict a density map similar  to ground truth in the positive part (\ie, $i$-th sample), while predicting an all-zero density map in the negative part (\ie, $j$-th sample). Note that we do not introduce category information during training \lw{since the probability of concatenating images with the same object type is negligible}.
%\NOTE{"dependent" here means what? The same object type in both images?}. 
The effectiveness of contrastive training without considering category information has been shown in various unsupervised and self-supervised learning methods~\cite{infonce, moco}.

\section{Experiments}

% In this section, we describe the implementation and experimental results in detail. 

{\bf Datasets.} We employ the \emph{FSC-147}~\cite{fsc147} for training and evaluating the proposed prompt counting model. It encompasses 147 distinct categories. 
%An important characteristic of FSC-147 is that the whole dataset is split into distinct train, validation, and test sets, where each set does not share any categories with other sets. This design ensures that objects of interest in the test set are not seen during the training process. 
Additionally, \emph{CARPK}~\cite{carpk} car counting dataset is used to assess the model's capability for cross-dataset adaptation. 
% It contains 1,448 images (89,777 cars) in parking lots from bird-eye view.

% \paragraph{Implementation Details}
% Regarding the \emph{network architecture}, we adopt ResNet-101~\cite{resnet} as the backbone for extracting \lw{image} features. Subsequently, the cross-attention module in transformer decoder~\cite{transformer} is employed to generate density features. Lastly, the density decoder from SPDCN~\cite{spdcn} is used to create density maps. The batch size is set to 12 and the resolution of all samples to $576\times 768$. Inputs undergo random color jitter, Gaussian blur, and horizontal flip. Additionally, random resizing is applied to each batch to enhance input scale robustness. Optimization is performed using the AdamW optimizer~\cite{adamw}. The basic learning rate is set at  $10^{-5}$, and cosine scheduler~\cite{sgdr} is applied to adjust learning rate dynamically. Moreover, we set $T = 2$ in (\ref{eq:fixpro}), meaning that the fixed-point function in (\ref{eq:fixpf}) undergoes two iterations during both training and inference.

\begin{table}[]
	\centering
	\scalebox{0.89}{
		\begin{tabular}{c|r|cc|cc}
			\hline
			\multirow{2}{*}{Prompt} & \multirow{2}{*}{Method} & \multicolumn{2}{c|}{validation}           & \multicolumn{2}{c}{test} \\
			&  & MAE & MSE & MAE & MSE  \\ \hline
			\multirow{10}{*}{box}   
			& GMN  		           & 29.66 & 89.81 & 26.52 & 124.57 \\
			& FamNet               & 26.55 & 77.01 & 26.76 & 110.95 \\
			& BMNet                & 17.89 & 61.12 & 16.89 & 96.65 \\
			& SPDCN 	           & 21.60 & 71.83 & 19.41 & 128.26 \\
			& CounTR	           & 17.40 & 70.33 & \textbf{14.12} & 108.81 \\
			& TFPOC                & --    & --    & 19.95 & 132.16 \\
			& ours                 & \textbf{16.87} & \textbf{59.45} & 16.68 & \textbf{105.08} \\ \cline{2-6}
			& SPDCN$^\dagger$      & 16.36 & 53.94 & 14.66 & 101.89 \\
			& CounTR$^\dagger$     & {13.15} & {49.72} & {12.06} & {90.01} \\ 
			& ours$^\dagger$	   & \textbf{12.80} & \textbf{48.65} & \textbf{11.86} & \textbf{89.40} \\ \hline
			\multirow{4}{*}{text}
			& ZSOC                 & 26.93 & 88.63 & 22.09 & 115.17 \\
			& CLIP-C               & 18.79 & 61.18 & 17.78 & 106.62 \\
			& TFPOC                & --    & --    & 24.79 & 137.15 \\
			& ours                 & \textbf{16.92} & \textbf{58.92} & \textbf{16.81}& \textbf{105.83} \\ \hline
			\multirow{2}{*}{point} 
			& TFPOC                & --    & --    & 20.10 & 132.83 \\
			& ours             	   & \textbf{17.16} & \textbf{59.38} & \textbf{15.86} & \textbf{103.27} \\ \hline
		\end{tabular}
	}
	\caption{Comparison with other prompt counting methods. Both box and point employ one annotation as the prompt. Withing box prompt, $\dagger$ means scale prior.}
	\label{tab:sota}
\end{table}

{\bf Evaluation metrics.} We assess performance using the metrics of Mean Absolute Error (MAE, $\texttt{MAE} = \tfrac{1}{N}\sum\nolimits_{i=1}^{N}|C_i - C'_i|$) and root Mean Squared Error (MSE, $\texttt{MSE} = \sqrt{\tfrac{1}{N}\sum\nolimits_{i=1}^{N}(C_i - C'_i)^2}$). In the formulation we use $N$ to represent the number of samples in either the validation or test set, $C_i$ and $C'_i$ correspond to the count of the $i$-th prediction and its corresponding ground truth. 
In ablation studies, we employ the average MAE/MSE across the three \lw{prompt types} (box/point/text) to conserve space. Comprehensive results can be found in the supplementary material.

\subsection{Prompt Counting Results}

% In this section, we present the experimental results of our prompt counting method and compare its performance against other class-agnostic counting methods. 
As shown in Table~\ref{tab:sota}, we conduct a comparative analysis across different prompt types, considering that fewer works have focused on combining these prompts together.

{\bf Box prompts:}
In the box-guided counting, our model demonstrates significantly lower MAE and MSE compared to similar simple-structured models, such as GMN~\cite{gmn} and FamNet~\cite{fsc147}. 
BMNet~\cite{bmnet} is a more complex network that incorporates a bi-linear matching module, and our model achieves results similar to it (MAE: 16.89 \textit{vs.} 16.68). For SPDCN~\cite{spdcn} and CounTR~\cite{countr}, we compare our model with them in two tracks, vanilla ones and models with scale prior($\dagger$). The scale prior, \textit{i.e.}, the width and height of box, is a specific property within box prompts. Specifically, SPDCN use it to adjust the receptive field, and CounTR design a test-time normalization (TT-norm) to refiene the prediction. With scale prior, SPDCN$^\dagger$ and CounTR$^\dagger$ achieves MAE of 16.36 and 13.15 on the validation set respectively. However, removing the scale prior from these models results in their MAEs increasing to 21.60 and 17.40 on the validation set (without $\dagger$ in Table~\ref{tab:sota}). To create a better comparison by accounting for the scale prior, we have implemented TT-norm, following CounTR's methodology.
By including the scale prior (ours$^\dagger$), our model's MAE significantly reduces to 12.80 and 11.86, which is better than SPDCN and CountTR with scale prior. 

Although the scale prior can boost the performance, we do not recommend implementing it since scale prior contradicts the motivation of unified prompt-based counting. It would be better to omits the scale prior and focuses on considering shared properties among prompts from different modalities (termed ``prompt masks'' in our paper) for prompt counting.

{\bf Text prompts:}
When compared with models trained using text prompts, the strengths of our model become evident. ZSOC~\cite{zsoc} utilizes a conditional VAE to generate class prototypes representing objects of interest, and then selects exemplars within a given image for counting. CLIP-C~\cite{clipcount} directly employs the frozen CLIP %~\cite{clip} 
model to generate density features. Subsequently, it designs a hierarchical text-patch interaction module to obtain density maps. While our model also leverages CLIP to transform text cues into visual features, we avoid fine-tuning CLIP's features. Instead, we use it to generate a prompt mask for consistent representation in prompt counting.  In the text-guided counting, our model achieves an MAE/MSE of 16.84/105.16, both of which are superior to ZSOC~(22.09/115.17) and CLIP-C~(17.78/106.62).

\begin{figure}[t]
	\centering
	\includegraphics[width=0.47\textwidth]{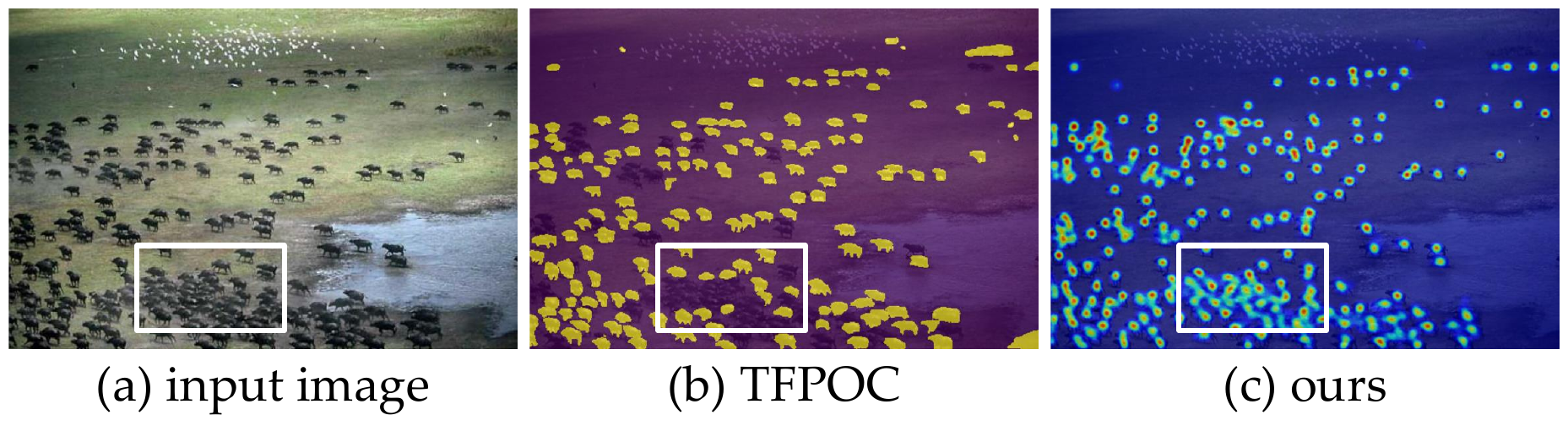}
	\caption{Visualization between TFPOC and our model. TFPOC cannot handle extremely dense regions (white box).}
	\label{fig:tfpoc}
\end{figure}

\begin{table}[t]
	\centering
	\scalebox{0.94}{
		\begin{tabular}{c|cc|cc|cc}
			\hline
			\multirow{2}{*}{Mask} & \multicolumn{2}{c|}{box~(val)}           & \multicolumn{2}{c|}{text~(val)}          & \multicolumn{2}{c}{point~(val)}         \\
			& {MAE}   & MSE   & {MAE}   & MSE   & {MAE}   & MSE   \\ \hline
			{cosine}         & {18.55} & 69.32 & {20.59} & 79.30 & {18.46} & 69.70 \\ 
			{softmax}        & \textbf{16.87} & \textbf{59.45} & \textbf{16.92} & \textbf{58.92} & \textbf{17.16} & \textbf{59.38} \\ \hline
		\end{tabular}
	}
	\caption{Comparison of text prompt mask generation.}
	\label{tab:clip}
\end{table}

{\bf Comparison with TFPOC:}
The concurrent approach TFPOC~\cite{tfpoc} is a training-free prompt counting model, leveraging SAM~\cite{sam} to count objects based on the segmentation results due to its high-quality zero-shot capability. It also incorporates box, text, and point as prompts facilitate object counting in the image. However, as demonstrated in Table~\ref{tab:sota}, our model outperforms TFPOC across all prompt types. This is attributed to our model's framework based on density prediction rather than instance segmentation, as the former excels in handling challenges like occlusion and blur, \abcn{while the latter may fail on small or densely placed objects.} Additionally, it is worth noting that our approach involves a training process, whereas TFPOC is entirely training-free. Figure~\ref{fig:tfpoc} provides an illustrative example showcasing our model's advantages, particularly in dense regions.

\abcn{Finally, in the supplemental, we provide results on using instance masks as prompts, as well as inference using multiple prompts at the same time.}

\subsection{Ablation Study on Text Prompt Mask}

Figure~\ref{fig:clip} illustrates that the text prompt mask generated via \texttt{cosine} distance method exhibits significant noise, whereas the \texttt{softmax} strategy effectively highlights the object of interest. In addition, we compare their respective counting performances on FSC-147, and Table~\ref{tab:clip} presents the MAE/MSE on the validation set, considering different approaches for generating text prompt masks. The results reveal that a noisy mask \abc{from the cosine strategy} not only increases estimation errors within its own track (text), but also degrades performance in other tracks (box \& point prompts).%\ANS{Here I retain the term "track" while using (text/box/point) to provide an explanation.}. 
The model has to contend with noisy input during training, potentially impacting its generalization capability.

\begin{table}[]
	\centering
	\begin{tabular}{c|cc|cc}
		\hline
		\multirow{2}{*}{loss function} & \multicolumn{2}{c|}{validation set}           & \multicolumn{2}{c}{test set} \\
		& MAE & MSE & MAE & MSE  \\ \hline
		L2    & 19.09 & 67.32 & 17.22 & 106.49 \\
		fixed-point & \textbf{16.98} & \textbf{59.25} & \textbf{16.45} & \textbf{104.72} \\ \hline
	\end{tabular}
	\caption{Comparison between MSE and fixed-point loss.}
	\label{tab:loss}
\end{table}

\begin{figure}[t]
	\centering
	\includegraphics[width=0.45\textwidth]{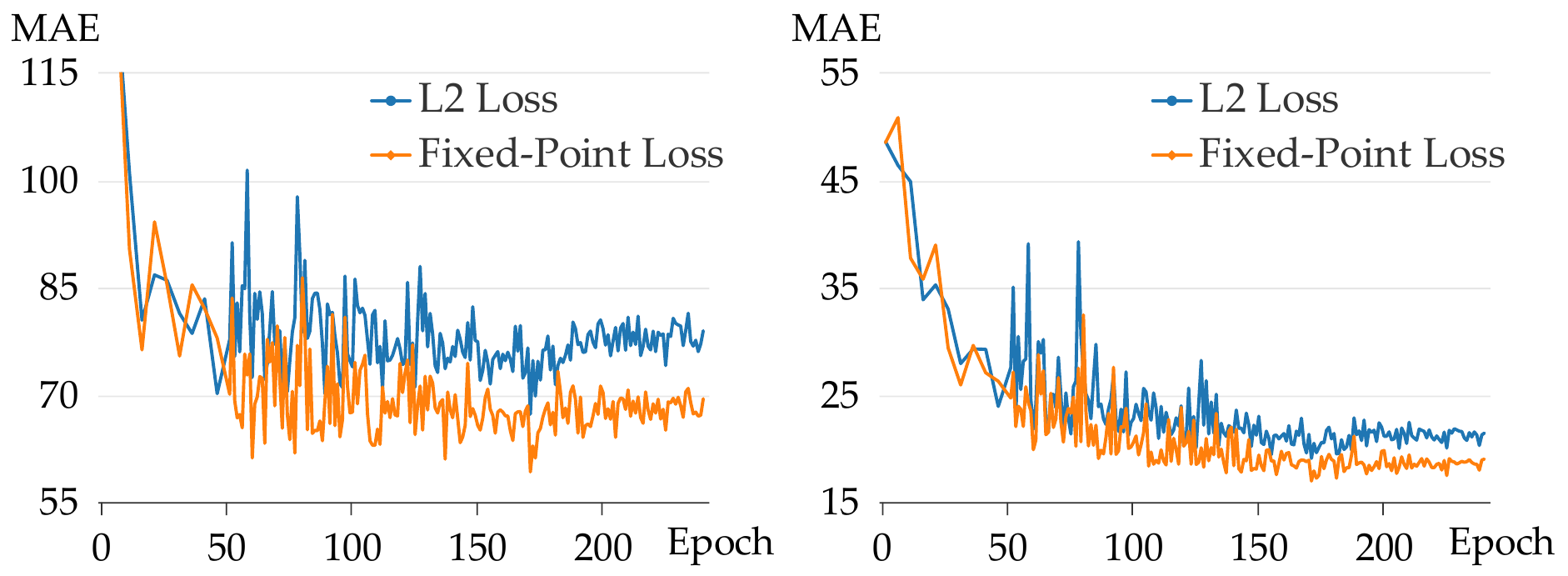}
	\caption{MAE/MSE on the validation set during training.}
	\label{fig:curve}
\end{figure}

\subsection{Ablation Study on Fixed-Point Loss}

{\bf Overall improvements.} We compare the proposed fixed-point loss in~(\ref{eq:floss}) with the vanilla L2 loss in~(\ref{eq:mse}) to demonstrate the overall improvements.
%, given that L2 is widely applied loss in counting tasks due to its ease of implementation and effectiveness. 
%However, 
As shown in Table~\ref{tab:loss}, when L2 loss is applied to prompt counting, the resulting MAE and MSE are 17.22 and 106.49, respectively. While this performance surpasses that of TFPOC~\cite{tfpoc}, the estimation errors are still considerably larger than the text-guided class-agnostic counting model, CLIP-C~\cite{clipcount}. In contrast, our fixed-point loss reduces the MAE (MSE) from 19.09 (67.32) to 16.98 (59.25) on the validation set, and the MAE (MSE) is lowered to 16.45 (104.72). Note that this improvement is achieved without introducing new network modules; rather, it involves looping the cross-attention and the density predictor while training with the designed fixed-point loss. Consequently, the number of parameters remains unchanged.

%\NOTE{what is V-MAE?}\ANS{I used V-MAE/MSE and T-MAE/MSE to differentiate the results on the validation set and test set, respectively. However, later on, I simply added a row to indicate the validation set and the test set, as shown in the first row in Table~\ref{tab:sota}}.
In Figure~\ref{fig:curve}, we also demonstrate the variation of \lw{MAE/MSE on the validation set} during training. The graph illustrates that the proposed fixed-point loss converges to a lower MAE/MSE than the vanilla L2 loss, making it a more effective choice compared to other loss functions.

\begin{table}[t]
	\centering
	\scalebox{1}{
		\begin{tabular}{cc|cc|cc}
			\hline
			\multirow{2}{*}{infinity} & \multirow{2}{*}{finity} & \multicolumn{2}{c|}{validation set}           & \multicolumn{2}{c}{test set} \\
			&						& MAE     	& MSE     	& MAE     	& MSE  \\ \hline
			$\mathcal{L}_{\infty}$	& --					& 24.91     & 71.89     & 26.97     & 110.21 \\ 
			--   					& $\mathcal{L}_{T'}$	& 20.79     & 68.73     & 17.38     & 107.41 \\
			$\mathcal{L}_{\infty}$ 	& $\mathcal{L}_{T'}$ 	& 17.14     & 62.14     & \textbf{16.29}     & 105.91 \\
			$\mathcal{L}_{\infty}$	& $\mathcal{L}_{T}$		& \textbf{16.98} 	& \textbf{59.25} 	& 16.45 	& \textbf{104.72} \\	\hline
		\end{tabular}
	}
	\caption{Ablation study on different combinations of infinity and finity part in fixed-point loss.}
	\label{tab:fxp}
\end{table}

\begin{figure}[t]
	\centering
	\includegraphics[width=0.48\textwidth]{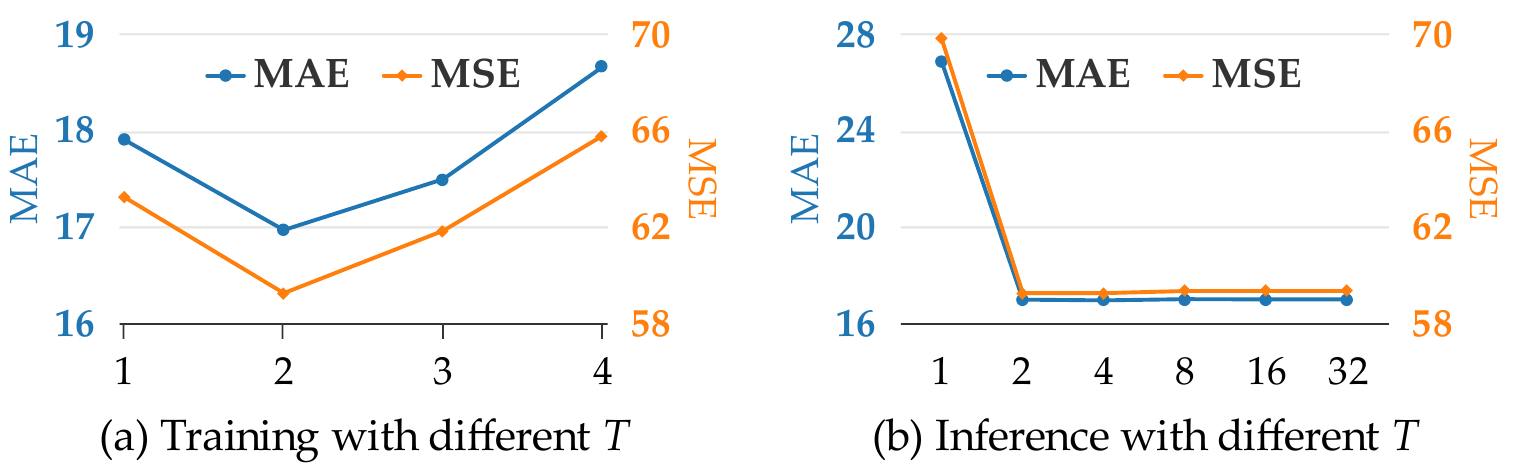}
	\caption{The influence of $T$ in training and inference.}
	\label{fig:loop}
\end{figure}

{\bf Detailed improvements.} We discuss three aspects of the proposed fixed point: the infinity loss $\mathcal{L}_{\infty}$ in (\ref{eq:lossi}), the finite loss $\mathcal{L}_{T}$ in (\ref{eq:losst}), and the number of iterations $T$. For the infinity loss, we can only conduct experiments on the model with or without it. Regarding the finite loss, there is an alternative approach which directly computes the L2 loss between $\bsb{d}^{(T)}$ and the ground truth $\bsb{d}'$, resulting in the loss:
\begin{align}
	\mathcal{L}_{T'} = \mathcal{L}(\bsb{d}^{(T)}, \bsb{d}'). \label{eq:losst2}
\end{align}
Note that this is different from the vanilla L2 loss (\ref{eq:mse}) since the recurrent structure is retained ($T = 2$). 

%\NOTE{what is T-MAE?}\ANS{See the answer in the last NOTE.}
Table~\ref{tab:fxp} presents a comparison of four combinations involving $\mathcal{L}_{\infty}$, $\mathcal{L}_{T}$ and $\mathcal{L}_{T'}$. When only loss $\mathcal{L}_{\infty}$ is used, the performance is poor (\lw{MAE} 26.97), and even worse than  TFPOC~\cite{tfpoc}. This outcome is reasonable since it completely removes the effect of prompts in the computation graph, failing to account for an incomplete prompt mask that differs from the ground truth. In the second experiment, $\mathcal{L}_{T'}$ in (\ref{eq:losst2}) serves as the learning objective. The results are significantly better than the previous approach, and its estimation errors on the test set are also lower than the L2 loss. By combining $\mathcal{L}_{\infty}$ and $\mathcal{L}_{T'}$, better MAE/MSE is achieved on the validation set, and the lowest MAE on the test set is achieved (16.29), but with slightly worse MSE than the proposed method(105.91 \textit{vs.} 104.72). 
%Besides, its performance on the validation set is inferior to the proposed combination.

\begin{table}[t]
	\centering
	\scalebox{0.9}{
		\begin{tabular}{c|cc|cc}
			\hline
			\multirow{2}{*}{\makecell{contrastive \\ training}} & \multicolumn{2}{c|}{validation set}  & \multicolumn{2}{c}{test set}    \\ 
			& {N-MAE}  & N-MSE  & {N-MAE}  & N-MSE  \\ \hline
			w/o   & 38.09 & 73.09 & 48.77 & 78.05 \\ 
			w/    & \textbf{1.90}  & \textbf{17.16} & \textbf{3.89}  & \textbf{15.19} \\ \hline
		\end{tabular}
	}
	\caption{Ablation study on contrastive training.}
	\label{tab:cts}
\end{table}

\begin{table}[t]
	\centering
	\scalebox{0.83}{
		\begin{tabular}{c|ccc|ccc}
			\hline
			\multirow{2}{*}{} & \multicolumn{3}{c|}{box}      & \multicolumn{3}{c}{text} \\ 
			&  BMNet & TFPOC & ours & CLIP-C   & TFPOC  & ours  \\ \hline
			MAE                 & 10.44 & 10.97 & \textbf{7.83} & 11.96    & 11.01  & \textbf{7.62}  \\ 
			MSE                 & 13.77 & 14.24 & \textbf{9.74} & 16.61    & 14.34  & \textbf{9.71}  \\ \hline
		\end{tabular}	
	}
	\caption{Cross-dataset adaptation on CARPK dataset.}
	\label{tab:carpk}
\end{table}

We next explore the number of iterations $T$ in both training and inference. We trained four models with $T \in \{1,2,3,4\}$ for comparison, as shown in Figure~\ref{fig:loop}(a). The best performance is obtained with $T=2$. A large $T$ causes the model to lose prompt mask information, while the model cannot reach its full effect without refinement ($T=1$). Focusing on the model trained with $T=2$, as shown in Figure~\ref{fig:loop}(b), we investigate whether applying different value of $T$ can reduce estimation errors during inference. It is noticeable that a significantly high MAE/MSE is observed when $T$ is set to 1, but the errors decrease and converge to the lowest value after the second iteration, providing evidence for the existence and validity of the fixed point in prompt counting. 
\subsection{Ablation Study on Contrastive Training}

To verify the effectiveness of contrastive training, we evaluate the trained model using negative samples. Similar to the training scheme, we extract a prompt token from one test sample, and use it to count on another randomly selected image to measure negative estimation errors, N-MAE and N-MSE, which are MAE and MSE for the negative samples with 0 ground-truth count.
%: $\texttt{N-MAE} = \frac{1}{N}\sum_{i=1}^{N}\|\hat{C}_{i}\|$ and $\texttt{N-MSE} = \sqrt{\frac{1}{N}\sum_{i=1}^N\hat{C}_i^2}$, since the ground-truth counts are 0 for negative samples. 
%. The ground truth counts are ignored for negative samples, as the count should be 0. 
Here we use category information to ensure that each pair of test samples contains different types of objects. Table~\ref{tab:cts} demonstrates that the contrastive training operates as expected, and without it, N-MAE/MSE are significantly higher. The errors are considerably reduced when the model is trained in contrastive training scheme.

\subsection{Cross-Dataset Adaptation}

Continuing the tradition of previous class-agnostic counting studies~\cite{fsc147, clipcount}, we explore cross-dataset adaptation by directly applying the prompt model to count cars in the CARPK dataset~\cite{carpk}. The results in Table~\ref{tab:carpk} provide a comparison between the proposed method and others on two prompt types: box and text prompts. Our method attains the lowest estimation errors in both tasks, showcasing its strong cross-dataset adaptation capability and the broad generalization of our prompt-based counting approach.

\section{Conclusion}

In this paper, we introduce a unified prompt-based counting model that can accurately count specific objects in an image using box, point, or text prompts. Our proposed method transforms prompts from different modalities into the a consistent representation, the prompt mask, which effectively highlights regions containing the objects of interest. Furthermore, we identify a fixed point within our framework when the predicted density map is treated as a prompt mask. This observation motivates us to implement a recurrent structure for refining the density map, and a fixed-point loss is also derived to make the training process stable and efficient. Addressing the inherent dataset bias, where most samples contain only one type of object in the current class-agnostic counting datasets, we employ a contrastive training shceme to mitigate shortcuts and bolster model robustness. Comprehensive experiments and ablation studies validate the effectiveness of our framework, demonstrating remarkable performance achievements.

\section{Acknowledgements}
This work was supported by a Strategic Research Grant from City University of Hong Kong (Project No. 7005665).

\bibliography{aaai24}

\end{document}